\pdfoutput=1
\documentclass[11pt]{article}

\usepackage{ACL2023}

\usepackage{times}
\usepackage{latexsym}

\usepackage[T1]{fontenc}
\usepackage[utf8]{inputenc}

\usepackage{microtype}

\usepackage{inconsolata}

\usepackage{spverbatim}

\usepackage{multirow}
\usepackage{booktabs}
\usepackage{bm}
\usepackage{wrapfig}
\usepackage{enumerate} 
\usepackage{amsmath}

\usepackage{hyperref}

\usepackage{graphicx}
\graphicspath{ {./images/} }

\usepackage{siunitx}

\title{Reranking for Natural Language Generation from Logical Forms: \\
A Study based on Large Language Models}

\author{Levon Haroutunian, 
    Zhuang Li, \\
    \textbf{Lucian Galescu,
    Philip Cohen,
    Raj Tumuluri,
    Gholamreza Haffari }
    \\
         Openstream.ai\\ 
         \texttt{\{levon, zhuang.li, lucian, phil.cohen, raj, reza.haffari\}@openstream.com}
         }

\begin{document}
\maketitle
\begin{abstract}

Large language models (LLMs) have demonstrated impressive capabilities in natural language generation. However, their output quality can be inconsistent, posing challenges for generating natural language from \emph{logical forms} (LFs). This task requires the generated outputs to embody the exact semantics of LFs, without missing any LF semantics or creating any hallucinations. In this work, we tackle this issue by proposing a novel generate-and-rerank approach. Our approach involves initially generating a set of candidate outputs by prompting an LLM and subsequently reranking them using a task-specific reranker model. In addition, we curate a manually collected dataset to evaluate the alignment between different ranking metrics and human judgements. The chosen ranking metrics are utilized to enhance the training and evaluation of the reranker model. By conducting extensive experiments on three diverse datasets, we demonstrate that the candidates selected by our reranker outperform those selected by baseline methods in terms of semantic consistency and fluency, as measured by three comprehensive metrics. Our findings provide strong evidence for the effectiveness of our approach in improving the quality of generated outputs.
\end{abstract}

\section{Introduction}
% motivate work
We consider the problem of natural language generation (NLG), which involves generating fluent and faithful utterances from structured meaning representations such as LFs~\cite{wang2021learning,chen-etal-2020-logic2text}. This task has gained significant importance, particularly for applications such as data augmentation for semantic parsing~\cite{wang2021learning} or question-answering systems~\cite{ribeiro2021investigating}, as well as response generation for dialogue systems~\cite{yu2019cosql}. This task plays a crucial role in enhancing the performance and capabilities of these systems by providing them with diverse and high-quality natural language utterances aligned with their underlying logical representations.

\begin{figure}
    \centering
    \includegraphics[scale=0.52]{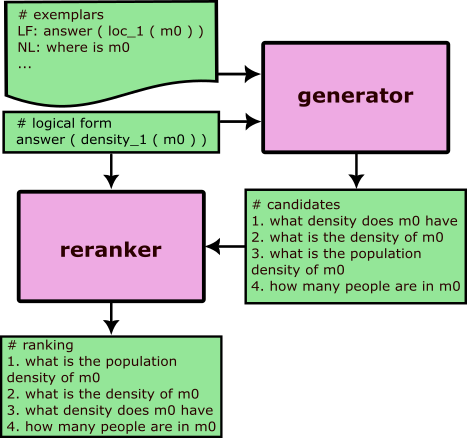}
    \caption{A high-level view of our approach. First, a generator model is given a set of exemplars and the LF of interest, from which it generates a set of candidates. The reranker is given this output, along with the LF, to produce a ranking of the candidates.}
    \label{fig:intro}
\end{figure}

LLMs have shown impressive performance across various NLG tasks~\cite{Chen2021EvaluatingLL,Ouyang2022TrainingLM}. However, the utterances generated based on LFs sometimes suffer from various deficiencies, such as hallucinations or missing parts of the input LF~\cite{chen-etal-2020-logic2text}. As depicted in Figure~\ref{fig:intro}, only 1 out of 4 candidates generated by the generator accurately and fluently reflects the semantic meaning of LF \texttt{answer(density\_1(m0))}. The remaining generated texts either introduce inaccuracies (\#4) or are awkwardly phrased (\#1 and \#2).

To improve the quality and fidelity of natural language generated from LFs, we take a generate-and-rerank approach that combines a fixed LLM generator with a finetuned reranker that discriminatively scores candidates given several pre-determined metrics~\cite{suzgun-etal-2022-prompt}. As in Figure~\ref{fig:intro}, our reranker successfully assigns the sole accurate and fluent candidate (\#3) generated by the generator a higher score than the other candidates. Furthermore, this method is very flexible: it can be applied to any dataset that pairs LFs with natural language, regardless of the formalism employed, and can be trained to align with any numeric metric.

While implementing our method, it became evident that a reliable reference ranking metric was necessary during both the training and evaluation phases of the reranker. However, determining the most suitable text quality evaluation measure for our specific task remained unclear. To address this, we manually curate an evaluation set, enabling us to thoroughly assess the alignment between various evaluation metrics and human judgement. By measuring the extent to which evaluation metrics accurately reflect human judgement, we are able to identify the most effective metrics for ranking the quality of generated texts and improve our generate-and-rerank approach.

Our contributions are:
\begin{itemize}
  \item We introduce a novel generate-and-rerank approach for generating natural language text from LFs using LLMs. This approach leverages the strengths of LLMs in initial text generation, followed by a reranking process to select the most fluent and semantically faithful candidates. The experiments show that our reranker significantly outperforms other candidate selection baselines across three datasets in terms of three evaluation metrics.

\item We conduct an in-depth analysis of various pre-trained metrics by utilizing a carefully curated dataset. This analysis allows us to identify and select the metrics that effectively produce rankings of natural language candidates, prioritizing fluency and semantic fidelity.

\item Through extensive experimentation, we provide valuable insights and recommend strategies for developing the optimal training data for a reranker, considering limitations on the generation budget. These strategies aim to maximize the performance and effectiveness of the reranking process.
\end{itemize}

\section{Related Work}\label{sec:lit}
% describe task of generating from LFs
\paragraph{NLG.} 
There is a large body of work concerning NLG from logical forms and/or structured data \citep{gardent-etal-2017-webnlg, chen-etal-2020-logic2text, parikh-etal-2020-totto, gehrmann-etal-2021-gem,shiri2022paraphrasing}. \citet{chen-etal-2020-logic2text} argues that NLG is best formulated as the task of generating text from LFs, as opposed to generating directly from structured data. This is the task of interest in our work, similar to others' work in SparQL-to-text \citep{Ngomo2013SorryID}, SQL-to-text \citep{xu-etal-2018-sql, Ma2021RelationAwareGT}, and AMR-to-text \citep{song-etal-2018-graph, Zhu2019ModelingGS, ribeiro-etal-2021-structural, Ribeiro2019EnhancingAG}.

Recent work considers the use of LLMs for few-shot NLG \citep{chen-etal-2020-shot, heidari-etal-2021-getting} and semantic parsing \citep{Drozdov2022CompositionalSP, shin-etal-2021-constrained, shin-van-durme-2022-shot,zhuo2023robustness} via in-context learning. Few-shot approaches to these tasks generally involve constructing a prompt containing a handful of training examples and sampling responses from an LLM without any training or fine-tuning beyond the LLM's pre-training. This method produces state-of-the-art results despite in some cases using only a fraction of the data required by other methods. Following these works, and specifically, the suggestion in \citet{shin-van-durme-2022-shot} that LLMs trained on code are suited to the task of semantic parsing because LFs are similar to code, we use Codex \citep{Chen2021EvaluatingLL} as our generation model.

\paragraph{Re-ranking.} 
This work is influenced by discriminative reranking approaches in machine translation \citep{lee-etal-2021-discriminative, bhattacharyya-etal-2021-energy}, semantic parsing \citep{Arcadinho2022T5QLTL}, abstractive summarization \citep{liu-liu-2021-simcls}, text generation \citep{langkilde-geary-2002-empirical, Deng2020Residual, li2022variational}, data-to-text \citep{harkous-etal-2020-text}, textual style transfer \citep{suzgun-etal-2022-prompt}, and mathematical reasoning \citep{cobbe2021training}.
 
\citet{lee-etal-2021-discriminative} introduces a discriminative reranking approach (DrNMT) for neural machine translation, utilizing a pre-trained language model to predict the BLEU score of a candidate translation given the source sentence. Unlike our approach, which employs a margin ranking loss function, they train DrNMT by minimizing the Kullback–Leibler divergence~\cite{kullback1951information} of the candidate and target scores.
Meanwhile, \citet{Arcadinho2022T5QLTL} employ a similar reranking approach in semantic parsing. Their T5QL model incorporates a ranking model (fine-tuned CodeBERT) to predict the correctness of a generated candidate parse from a given natural language question. In contrast, our model uses a similar architecture but works in reverse, generating text from LFs.
\citet{liu-liu-2021-simcls} present a contrastive learning method, SimCLS, for ranking abstractive summarization candidates. The authors finetune a RoBERTa encoder to measure the alignment of a summary with the text it originates from: the embedding of a higher quality summary will be more similar to the embedding of the original text than the embedding of a lower quality summary. Similar to our work, they train their model by minimizing a ranking loss function.

\section{Reranking Approach}\label{sec:method}

In this section, we present details of our methods, including our choice of generator model, reranker architecture, and  evaluation metric.

\subsection{Problem Formulation}
Given a pool of LFs paired with their natural language utterances, our task is to generate a natural language utterance $y$ corresponding to an LF $x$. In this work, we first generate a set of $n$-best candidates $\hat{\mathcal{Y}}_x :=  \{ \hat{y}_1, \hat{y}_2, ..., \hat{y}_n \}$, and then rerank them using a reranker based on a \emph{quality score}. We assume that with access to one ground truth utterance $y$ corresponding to the input LF $x$, we would be able to calculate the quality score for each candidate using a function $Q( \hat{y}_i | x, y )$. In our setting, $Q$ is an automatic metric to score the quality of a generated candidate text against the ground-truth text, such as BLEU~\citep{papineni-etal-2002-bleu}. These quality scores would determine the relative ranking of the $n$-best candidates, and would allow us to choose the optimal text output. 

Our goal is to train a reranker model to predict the relative order of the values assigned by $Q$ given only $x$; that is, without access to the gold reference $y$. This is achieved by training the parameters $\theta$ of the scoring function $R_{\theta}(\hat{y}_i | x)$.

\subsection{Generator}
\label{sec:cand_gen}
We prompt Codex \citep{Chen2021EvaluatingLL} in a few-shot setting to generate natural language candidates for a given LF. Each prompt includes a number of exemplars\footnote{It is 15 in our experiments. } randomly drawn from the training set, presented as simple input/output pairs. An example prompt is given in Appendix \ref{sec:codex_prompt}.

To create training data for the reranker model, we generate natural language candidates for LFs in the training set by repeatedly prompting Codex\footnote{We use the \texttt{code-davinci-002} model of Codex, which has around 175B parameters, with a temperature of 0.7 in our experiments.} until there are $n$ unique candidates per logical form. At inference time, we construct prompts for each LF in the test set in much the same manner. The score $G(\hat{y}|x)$ denotes the log-probability of $\hat{y}$ given the input $x$ by Codex. 

\subsection{Reranker}\label{sec:reranker}

Our reranker model is composed of CodeBERT \citep{feng-etal-2020-codebert} as the base model and a feed-forward regression head over the \texttt{[CLS]} token. 

For each forward pass, the input to the reranker consists of a LF concatenated with a natural language candidate, separated with an EOS token. The output is a single real-valued number that represents the relative quality of the candidate.

We finetune the model using the Huggingface library\footnote{\href{https://huggingface.co/}{huggingface.co}}. We also use the publicly available checkpoint for CodeBERT (\texttt{microsoft/codebert-base})\footnote{\href{https://github.com/microsoft/CodeBERT}{github.com/microsoft/CodeBERT}}, which has approximately 110M parameters.

\paragraph{Loss Function}
The training objective for our reranker is to minimize a weighted margin ranking loss across pairs of natural language candidates. For each set of candidates corresponding to one LF, the loss is,
$$
    \label{eq:loss}
    L (\theta) = \dfrac{
    \sum_{i, j; i \neq j}^n 
    \max[0, - z_{i,j}  ( \hat{z}_{i,j} + \gamma ) ]   }{ n (n-1) } 
$$
where  $n$ is the number of candidates, and $\gamma$ represents a margin. The value of $z_{i,j} := Q(\hat{y}_i|x,y) - Q(\hat{y}_j|x,y)$  is the difference between the gold quality scores of candidates $i$ and $j$. Its magnitude reflects the relative importance of obtaining the correct ranking for the pair. The score $\hat{z}_{i,j} := R_{\theta}(\hat{y}_i|x) - R_{\theta}(\hat{y}_j|x)$ represents the predicted difference between candidates $i$ and $j$. 

\subsection{Scoring of the Candidates}

At test time, we use either the re-ranker score or its combination with the generator probability score to select the winning candidate in the $n$-best list. The combined score is

\begin{equation}
    \lambda R_{\theta}( \hat{y}_i|x) + (1- \lambda) G( \hat{y}_i|x) 
\end{equation}
where  $\lambda$ is a hyperparameter that is tuned on the development set. In practice, we found that the value of $\lambda$ did not generalize well across datasets or even across seeds; a separate $\lambda$ value was thus tuned for each model run.

\section{Evaluation of Text Generation Metrics for Reranking}\label{sec:metric}

Automatic evaluation of (generated) text quality is not an easy task, and poses challenges for building the reference rankings for candidate sets in the training set and fairly evaluating the generation ability of generators. Therefore, we curate an evaluation set to evaluate the effectiveness of several text generation metrics. 

\paragraph{Generation Metrics.} We consider the following pre-existing metrics\footnote{We do not finetune or otherwise modify these metrics.}:

\begin{itemize}
\item BLEU\footnote{We use the NLTK implementation of BLEU; \href{https://www.nltk.org/}{nltk.org}} \citep{papineni-etal-2002-bleu}, which explicitly measures the lexical overlap between reference and hypothesis. 

\item BERTScore \citep{Zhang2019BERTScoreET} and BLEURT \citep{sellam-etal-2020-bleurt, pu-etal-2021-learning}, which frame evaluation as a regression task. 

\item PRISM \citep{thompson-post-2020-automatic} and BARTScore \citep{Yuan2021BARTScoreEG}, which frame evaluation as a generation task. 
\end{itemize}

We also use probability scores from a semantic parser as an additional metric. For each dataset, we train a semantic parser by finetuning CodeT5 \citep{wang-etal-2021-codet5} to generate LFs from natural language utterances. Then, we use the trained parser to calculate the probability that a generated candidate is parsed to the original LF. This score measures the faithfulness of the candidate to the original semantics of the LF.  Unlike other metrics above, the parser probability is calculated based on the generated candidate and the LF, as opposed to the generated candidate and the reference text.

In addition to evaluating individual metrics, we also evaluate their combinations. When combining metrics, we first normalize the scores for each metric so that the mean score is 0 and the standard deviation is 1, and then we sum the normalized scores across possible metrics. We normalize the scores in order to ensure that each metric is given the same weight in relation to the others.

\paragraph{Curation of Evaluation Data.} To determine the alignment of each metric with human preferences, we constructed a small, manually-crafted evaluation set. We randomly selected 200 LFs from the train split of CFQ-MCD1 \citep{Keysers2019MeasuringCG}, each with eight generated candidates. Each candidate is labelled either `correct' or `incorrect'; rather than producing a strict ranking in this evaluation set, we instead opted for binary classes to allow for the fact that multiple candidates can be equally acceptable. 

We developed a set of criteria to account for both semantic accuracy and fluency, presented below:
\begin{enumerate}[(i)]
    \item If a candidate \textit{omits} a piece of information that appears in the reference, it is incorrect.
    
    \item If a candidate \textit{inserts} or \textit{substitutes} a piece of information that does not appear in the reference, it is incorrect.

    \item If a candidate is markedly \textit{less fluent} (e.g. contains unnatural constructions) compared to other candidates in the set, it is incorrect.

    \item If a candidate contains terms that appear in the LF (e.g. \texttt{?x0}) but should not appear in the utterance, it is incorrect.

    \item Otherwise, the candidate is correct.
\end{enumerate}
Using these criteria, a human annotator assigned binary labels to the candidates in the evaluation set.

\paragraph{Evaluation Measures.} To assess the alignment between the chosen metrics and human judgements, we calculated (i) top-1 accuracy, or the probability that the highest-scoring candidate in a set belongs to the `correct' class; and (ii) ranking accuracy, or the probability that any `correct' candidate is ranked above any `incorrect' candidate. In the calculation of these values, the sets of candidates in the evaluation set that are comprised of only one class (i.e., either all are incorrect or all are correct)  are excluded.

\paragraph{Results.} Table \ref{tab:metrics} provides the results of our evaluation for each metric, as well as for the combination of all metrics and the best-performing combination (BLEURT + PRISM + Parser).

Our findings support the conclusion by \citet{WMT22} that trained metrics outperform BLEU, and the suggestion by \citet{Amrhein2022ACESTA} and \citet{Moghe2022ExtrinsicEO} that a combination of different families of metrics is likely to be stronger than any one metric alone. Each of the three metrics in our best-performing combination work in very different ways: BLEURT is an encoder-only model that is trained to predict direct assessment scores assigned to machine translation outputs by human evaluators \citep{pu-etal-2021-learning}, PRISM is an encoder-decoder model trained for NMT and deployed as a zero-shot paraphraser \citep{thompson-post-2020-automatic}, while our parser is an encoder-decoder model trained to convert text into LFs. We suspect that the differences between these models contribute to their strength in combination. Furthermore, we speculate that parser probability scores reflect the semantic consistency between a candidate and the reference LF, while BLEURT and PRISM scores more strongly reflect a candidate's surface-level similarity to the reference text and its overall fluency.

Following these results, we use the combination of scores assigned by BLEURT, PRISM, and the task-specific semantic parser to determine reference rankings for training the reranker.

%%%%%%%%%%%
\begin{table}[t]
    \centering
    \small
    \begin{tabular}{|l|c|c|}
       \hline
       \textbf{Metric}  & \textbf{Top-1} &  \textbf{Ranking} \\
       \hline
        BARTScore & $70.62$ & $75.57$ \\
        BERTScore & $73.45$ & $78.11$ \\
        BLEU & $66.67$ & $73.40$ \\
        BLEURT & $71.75$ & $81.96$ \\
        PRISM & $76.27$ & $81.19$ \\
        Parser & $76.27$ & $79.41$ \\
        Combination of above metrics & $79.10$ & $82.59$ \\
        BLEURT + PRISM + Parser & \bm{$81.36$} & \bm{$84.22$} \\ 
    \hline
    \end{tabular}
    \caption{Top-1 accuracies and ranking accuracies (represented as percentages) across the evaluation dataset.}
    \label{tab:metrics}
\end{table}
%%%%%%%%%%%%

\section{Experiments}\label{sec:experiments}

\subsection{Datasets}\label{sec:data}
We conducted experiments using three datasets:
\begin{itemize}
    \item \textbf{GeoQuery:} This dataset consists of 880 English questions focusing on the geography of the United States \citep{Zelle1996LearningTP}. We report results for both the standard split and the query split \cite{finegan2018improving}. The train and test sets in the query split contain distinct sets of LFs.
    \item \textbf{Jobs:} The Jobs dataset comprises 640 English queries that correspond to LFs in a jobs database \citep{Califf1999RelationalLO}. We present results based on the standard split of this database.
    \item \textbf{CFQ:} CFQ contains approximately 239,000 synthetic English questions paired with SPARQL queries \citep{Keysers2019MeasuringCG}, with three different data splits designed to maximize compound divergence between training and test sets. Our study focuses on the MCD1 split, which consists of 96,000 training pairs and 12,000 test pairs.
\end{itemize}

For each dataset, we generate natural language candidates for LFs in the training set as described in section \ref{sec:cand_gen}, with $n = 8$ natural language candidates per logical form. The candidate generation process requires repeated calls to Codex, which is a significant bottleneck. Consequently, only 30k training pairs (240k total candidates) are used for experiments on CFQ-MCD1. Following \citet{Drozdov2022CompositionalSP}, we map Freebase identifiers to simpler keywords. See Appendix \ref{sec:mapping} for the full mapping.

\subsection{Baselines}

We compare the performance of our model against three different baselines and one ORACLE method.

\begin{itemize}
    \item \textbf{Random selection:} A candidate is selected randomly from the set of unique candidates generated for each LF. This method serves as a lower bound.

    \item \textbf{Self-consistency:} The most frequently appearing candidate is selected. Ties are broken randomly. This method is proposed in \citet{Wang2022SelfConsistencyIC} for use with chain-of-thought style prompting and is extended for use with simpler prompting styles in \citet{Drozdov2022CompositionalSP}. 

    \item \textbf{Highest generator probability:} For each candidate, the token log probabilities given by the generator are averaged. The selected candidate is the one with the highest score.

    \item \textbf{Oracle:} Scores for each of the three metrics (BLEURT, PRISM, and parser probability) are normalized and summed for each candidate. The candidate with the highest combined score is selected. The performance of this method serves as an upper bound.
\end{itemize}

\begin{table*}
    \centering
    \small
    \begin{tabular}{|l||ccc|ccc|ccc|ccc|}
        \hline
        \multirow{2}{*}{\textbf{Method}} &
        \multicolumn{3}{c|}{Jobs} & \multicolumn{3}{c|}{GeoQuery (standard) }
        & \multicolumn{3}{c|}{GeoQuery (query)} &
        \multicolumn{3}{c|}{CFQ-MCD1} 
        \\
        {} & 
        bleurt & parser & prism &
        bleurt & parser & prism &
        bleurt & parser & prism &
        bleurt & parser & prism  \\
        \hline
        Oracle & 
        $70.7$ & $95.0$ & $6.9$ &
        $86.0$ & $93.6$ & $40.3$ &
        $86.1$ & $89.3$ & $40.3$ &
        $71.5$ & $64.3$ & $22.2$ \\
        Random & 
        $59.4$ & $84.7$ & $3.0$ &
        $74.2$ & $77.6$ & $23.1$ &
        $74.4$ & $85.2$ & $22.7$ &
        $61.1$ & $44.7$ & $13.0$\\
        Self-cons. & 
        $60.7$ & $89.1$ & $3.4$ &
        $77.5$ & $79.7$ & $26.5$ &
        $78.3$ & $85.2$ & $27.2$ &
        $62.7$ & $47.1$ & $14.3$\\
        Generator &
        $61.3$ & $93.0$ & $3.7$ &
        $77.8$ & $81.8$ & $28.1$ &
        $79.0$ & $85.7$ & $29.0$ &
        $64.6$ & $49.0$ & $15.8$\\
        Reranker & 
        \underline{\bm{$62.7$}} & \underline{$93.5$} & \underline{\bm{$4.4$}} &
        \underline{$81.0$} & \underline{$90.9$} & \underline{$32.5$} &
        \underline{$79.9$} & \underline{\bm{$89.8$}} & $28.9$ &
        \underline{$66.3$} & \underline{\bm{$61.9$}} & \underline{$17.2$}\\
        Combined & 
        \underline{\bm{$62.7$}} & \underline{\bm{$93.7$}} & \underline{$4.3$} &
        \underline{\bm{$81.2$}} & \underline{\bm{$91.1$}} & \underline{\bm{$32.9$}} &
        \underline{\bm{$80.0$}} & \underline{$89.7$} & \bm{$29.1$} &
        \underline{\bm{$66.5$}} & \underline{\bm{$61.9$}} & \underline{\bm{$17.4$}}\\
        \hline
         
    \end{tabular}
    \caption{BLEURT, parser, and PRISM scores for re-ranker and baselines. The oracle method selects the candidate with the highest combined BLEURT, PRISM, and parser score; the random method selects a candidate at  random; the self-consistency method selects the most frequently generated candidate; the generator method selects the candidate with the highest probability from the generator (conditioned on the prompt); the reranker method selects the candidate with the highest score from the trained reranker; and the combined method selects candidates based on a linear combination of generator and reranker scores. Parser and PRISM scores are probabilities represented as percentages. Bold values represent the highest (non-oracle) score in the column, and underlined scores represent a statistically significant improvement from generator scores (\textit{p} < 0.01).}
    \label{tab:combined_metrics}
\end{table*}

\subsection{Training Details}

At the beginning of training, the base model of the reranker is frozen, and loss is only backpropagated through the regression head. After 10 epochs, the final layer of the base model is unfrozen, and loss is backpropagated through both the regression head and this final layer of CodeBERT for the remainder of the training. The optimization details are given in Appendix~\ref{sec:details}.  

\subsection{Main Results}\label{sec:results}

Our experiment results can be seen in Table~\ref{tab:combined_metrics}, depicting the performance of our reranker model. Training of the reranker is performed five times with different seeds, and we report the mean score. 

Importantly, the reranker significantly outperforms the generator baseline regarding parser probability. Specifically, there is an impressive absolute difference of up to 12.9 percentage points (for CFQ-MCD1). The reranker also shows modest gains over the generator baseline in PRISM and BLEURT scores. These metrics suggest that the reranker's selected candidates have both greater semantic consistency and slightly enhanced fluency than those chosen without reranking. 

It is worth noting that the performance gap between the reranker and the highest probability baseline is most prominent in the GeoQuery standard split. In contrast, the other three datasets were deliberately designed to assess models' compositional generalization capabilities. For instance, both the query splits of GeoQuery and the Jobs datasets have no LF overlap among their train and test sets, and CFQ-MCD1 was split to maximize the compound divergence between the two sets. The smaller performance gap between the generator and the reranker on these three datasets suggests that the generator model, Codex, has stronger compositional generalization abilities than the finetuned reranker. 

Additionally, we observed that the self-consistency method performs poorly compared to other baselines, such as candidate selection based on generator probability. This finding indicates that self-consistency is not a helpful selection method for this particular task.

\subsection{Influence of the Size of the $n$-Best List}

\begin{figure}
    \centering
    \includegraphics[scale=0.7]{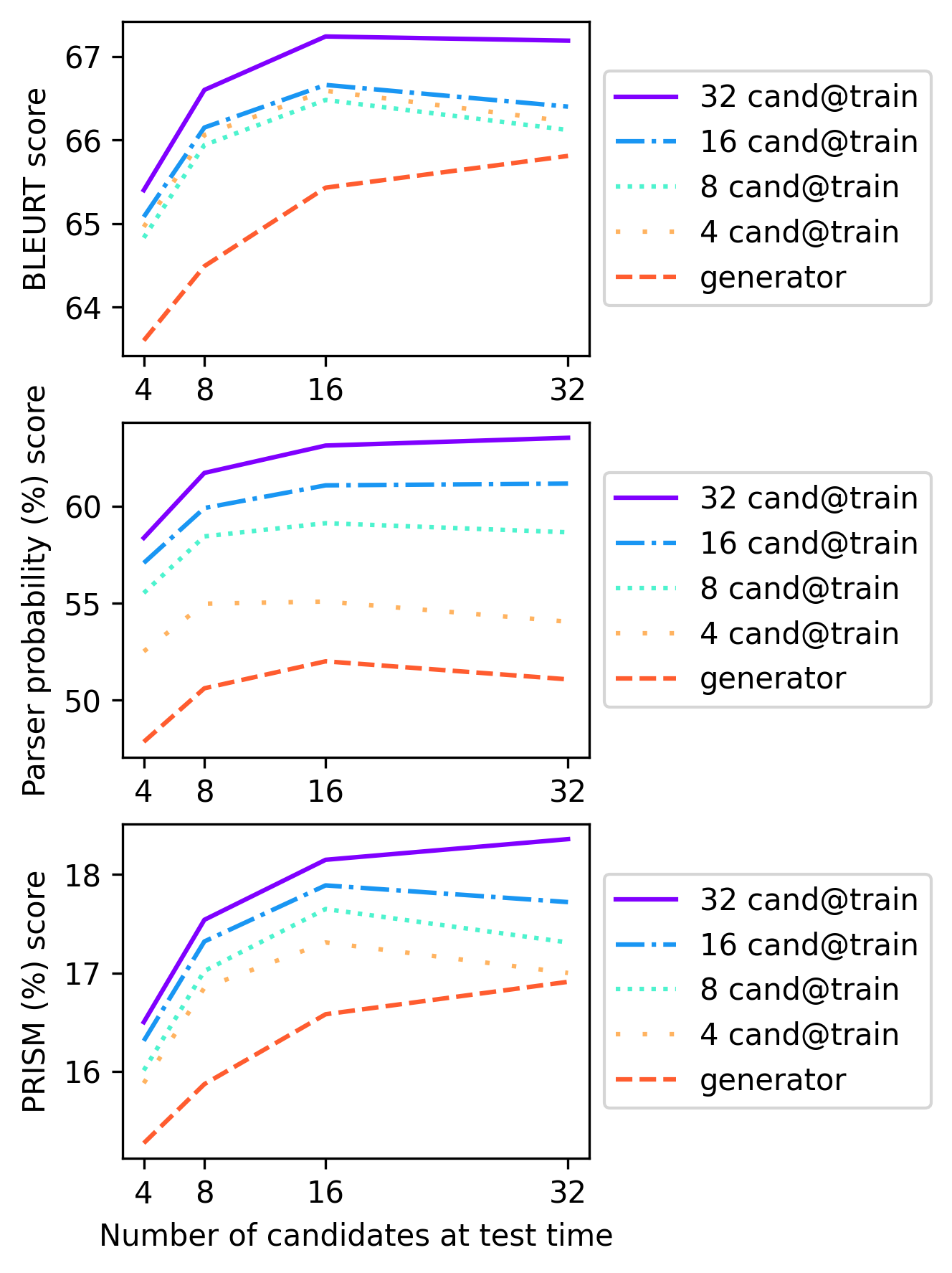}
    \caption{BLEURT, parser probability, and PRISM scores across different sizes of candidate lists seen at training and test time.}
    \label{fig:n-best}
    \vspace{-3mm}
\end{figure}

In this experiment, we examine the effect of different sizes of candidate lists seen during train and test time. We use a subset of the CFQ-MCD1 dataset in these experiments due to time constraints; specifically, we randomly select 3,000 data pairs from the train split (further divided into 2,700 train pairs and 300 dev pairs) and report our results on 1,200 randomly selected data pairs from the test split. The results of this experiment are shown in Figure \ref{fig:n-best}.

While scores for all metrics improve as the train-time $n$-best list grows, the most significant gains are observed in parser probability. This suggests that increasing the number of candidates per LF that the reranker sees at training time is an effective way to increase the semantic consistency of candidates chosen by the reranker. The performance on BLEURT and PRISM increases more slowly as the number of candidates seen at train time increases, with the largest increase happening in the jump from 16 candidates to 32, suggesting that it may be necessary to generate many more candidates per LF in order to substantially improve the fluency of selected candidates. Additionally, increasing the number of candidates seen at \textit{test} time appears to have a negligible effect on the semantic consistency of candidates selected, but a notable effect on the BLEURT and PRISM scores. Scores for these metrics increase steadily for the first three sizes of candidate lists at test time, regardless of the number of candidates seen at train time. 

However, performance on these metrics drops when the test size increases to 32 candidates for rerankers trained with a candidate list size of 16 and below. We hypothesize that these observations are due to changes in the quality and diversity of the test candidates. As the number of candidates per LF increases, it is more likely that any given candidate set will contain high-quality candidates. Increasing the candidate set size also increases the diversity of candidate sets. Improvements in test candidate set quality appear to be helpful for sizes up to $n = 16$, but models trained on smaller candidate sets may not be able to generalize well to candidate set sizes of $n = 32$ due to the larger degree of diversity. However, the reranker trained with 32 candidates per LF is able to take advantage of further quality improvements in the largest test candidate list size due to its exposure to diverse candidate sets.

\subsection{Fixed Generation Budget}

As generating large-sized $n$-best lists from Codex is time-consuming, we consider a scenario in which the time budget for training data generation is fixed. When given limited time to generate training data, is it better to prioritize coverage of as many LFs as possible by considering small $n$-best lists, or is it better to ensure that there is a large number of candidates for each LF in the training set? 

We generate natural language candidates for LFs in the training set of CFQ-MCD1 over one 24-hour period. We use Codex to generate 4, 8, 16, or 32 candidates per LF for 24 hours. We also generate a dataset containing a \emph{variable} number of candidates per LF. To do this, we prompt Codex for 10 candidates per LF, then discard duplicates. Any LFs with candidate sets of length 1 are also discarded. This results in a dataset that pairs LFs with sets of candidates with a minimum length of 2 and a maximum length of 10. The average number of candidates per LF in this dataset is 7.6 in our trial; see  Table \ref{tab:data-sizes}.

A reranker is then trained for each dataset using the method described in Section \ref{sec:reranker}.\footnote{For the dataset with a variable number of candidates per LF, the loss for each candidate set is multiplied by a weight term, which is calculated as the size of the candidate set divided by the average candidate set size. This is done in order to normalize the magnitude of gradient updates across the training set. } Each reranker is evaluated on the full test split of CFQ-MCD1, with eight candidates per LF.

The results  are shown in Table \ref{tab:fixed-budget}. The reranker trained on the dataset with a variable number of candidates per LF has the best performance as measured by BLEURT and PRISM, and the second best performance as measured by parser probability. Its strong performance is likely due to the fact that it is trained on the largest dataset (at 93k total candidates) that also covers the largest number of LFs (12k).  This  suggests that the best way to use a limited budget for generating reranker training data is to maximize the total quantity of generated candidates; ensuring a large (or even consistent) number of candidates per LF is less important. Using a variable candidate set size is also more efficient in a pay-per-token setting, as fewer duplicated candidates will be discarded than there would be with a fixed candidate set size.

\begin{table}[t]
    \centering
    \begin{tabular}{|c|c|c|}
        \hline
       Cands. per LF  & Total LFs & Total cands. \\
       \hline
        4 & 11,395 & 45,580 \\
        8 & 8,147 & 65,156\\
        16 & 4,076 & 65,216\\
        32 & 1,935 & 61,920\\
        Variable & 12,253 & 93,226 \\
        \hline
    \end{tabular}
    \caption{Resulting dataset sizes after generating a specified number of natural language candidates per LF.}
    \label{tab:data-sizes}
\end{table}

\begin{table}[t]
\resizebox{\columnwidth}{!}{
    \centering
    \begin{tabular}{|l|ccc|}
        \hline
        \textbf{Set size} & bleurt & parser (\%) & prism (\%)  \\
        \hline
        4 & $66.1 \pm 0.4$ & $57.5 \pm 1.1$ & $17.0 \pm 0.4$ \\ 
        8 & $65.8 \pm 0.2$ & $60.4 \pm 0.9$ & $16.8 \pm 0.1$ \\ 
        16 & $65.7 \pm 0.2$ & \bm{$60.8 \pm 1.1$} & $16.8 \pm 0.2$ \\ 
        32 & $66.0 \pm 0.3$ & $59.7 \pm 0.3$ & $16.9 \pm 0.2$ \\ 
        Variable & \bm{$66.2 \pm 0.2$} & $60.4 \pm 0.6$ & \bm{$17.1 \pm 0.1$} \\ 
        \hline
        Oracle & $71.5$ & $64.3$ & $22.2$ \\
        Generator & $64.6$ & $49.0$ & $15.8$ \\ 
        \hline
    \end{tabular}
}
\caption{Scores for different candidate generation configurations, given 24 hours of generation time with Codex. Set size refers to the number of candidates per LF in the training set. The oracle method selects the candidate with the highest sum of normalized BLEURT, parser, and PRISM scores. The generator method selects the candidate with the highest probability assigned by Codex.}
    \label{tab:fixed-budget}
\end{table}

\subsection{Using Instruction-Following LLMs}

We perform the next experiment in order to determine the effectiveness of a general-purpose language model in the role of generator in place of a language model optimized for code, or in the role of reranker in place of a discriminative model. We generate eight candidates per LF by prompting either Codex (as in previous experiments) or ChatGPT. Then, we rerank the candidates either using a finetuned reranker (as in previous experiments), or  by prompting ChatGPT. We replicate our experiments on the GeoQuery dataset using GPT-3.5-turbo (ChatGPT\footnote{\href{ https://platform.openai.com/docs/model-index-for-researchers}{platform.openai.com/docs/model-index-for-researchers}}) \citep{Ouyang2022TrainingLM}. The details of generation from ChatGPT are given in Appendix \ref{sec:chatgpt}. 

The results\footnote{ The scores reported for the two configurations using CodeBERT as the reranker represent the mean score over five trials, while the trials using GPT-3.5 report the score from one trial.} are presented in Table \ref{tab:chatgpt}. The best-performing combination by a wide margin is Codex as the generator, and fine-tuned CodeBERT as the reranker. Using ChatGPT does not appear to add benefits for either the candidate generation step or the reranking step. We performed manual error analysis to determine why the performance gap was so wide, the results of which are presented below.

\begin{table}[t]
\small
    \centering
    \begin{tabular}{|l|l||ccc|}
    \hline
        Generator & Reranker & bleurt & parser & prism  \\
        \hline
        Codex & CodeBERT & \bm{$79.9$} & \bm{$89.8$} & \bm{$28.9$} \\
        Codex & GPT-3.5 & $76.9$ & $86.9$ & $25.3$   \\
        GPT-3.5 & CodeBERT & $71.0$ & $71.9$ & $8.2$ \\
        GPT-3.5 & GPT-3.5 & $71.9$ & $75.6$ & $8.5$ \\
        \hline
    \end{tabular}
    \caption{Scores for different combinations of models used as generators and rerankers. GPT-3.5 here refers to OpenAI's ChatGPT.}
    \label{tab:chatgpt}
\end{table}

\paragraph{Problems with Generation.}

The style of candidates generated by Codex tended to more closely match the style of the gold natural language utterances than did the candidates generated by ChatGPT. Namely, candidates generated by ChatGPT tended to use sentence casing and end punctuation, while Codex candidates tended to be all lowercase with no punctuation, as are the GeoQuery questions. Additionally, the ChatGPT candidates used more varied language than the Codex candidates did. While these surface-level differences may not reflect a difference in candidate \textit{quality}, it is possible that they are penalized by automatic metrics such as the ones we use here.
A more concerning finding is that the candidates generated by ChatGPT tended to include more frequent and more severe hallucinations than those generated by Codex; see Table \ref{tab:chatgpt_candidates} for examples.

We speculate that these differences in generated candidates are due to the fact that Codex is directly optimized for tasks that involve code, which makes it a better fit for the task of generating text from structured meaning representations. While ChatGPT's training does include tasks involving code, many of its training tasks do not concern code.

\begin{table}[t]
\small
    \centering
    \begin{tabular}{|c|c|}
 \hline
        \multicolumn{2}{|c|}{\textbf{Codex Candidates}} \\
        \hline
        \multicolumn{2}{|l|}{what states are in the usa} \\
        \multicolumn{2}{|l|}{what are the names of the states} \\
        \multicolumn{2}{|l|}{what state is this} \\
        \multicolumn{2}{|l|}{what are all the states} \\
        \multicolumn{2}{|l|}{what states are there} \\
        \multicolumn{2}{|l|}{name all the states} \\
        \multicolumn{2}{|l|}{what states exist} \\
        \multicolumn{2}{|l|}{what are the states} \\
        \hline \hline
        \multicolumn{2}{|c|}{\textbf{ChatGPT Candidates}} \\
        \hline
        \multicolumn{2}{|l|}{Which states are in the United States?} \\
        \multicolumn{2}{|l|}{What are the names of all the states?} \\
        \multicolumn{2}{|l|}{How many states are there in the country?} \\
        \multicolumn{2}{|l|}{Which states are located in the Midwest?} \\
        \multicolumn{2}{|l|}{What states make up New England?} \\
        \multicolumn{2}{|l|}{What is the largest state in terms of land area?} \\
        \multicolumn{2}{|l|}{Which states have coastline?} \\
        \multicolumn{2}{|l|}{What is the capital of each state?} \\
    \hline
       \hline
      \textbf{Logical Form:}  & answer ( state ) \\
      \hline
       \textbf{Gold Utterance:} & list the states \\
       \hline
    \end{tabular}
    \caption{A comparison of candidates generated by Codex and ChatGPT. While the candidates generated by Codex are faithful to the style of the gold question and are mostly semantically consistent with the given LF, the candidates generated by ChatGPT include substantial hallucinations.}
    \label{tab:chatgpt_candidates}
\end{table}

\paragraph{Problems with Reranking.}

When using ChatGPT as a reranker, we found that it returned a natural language sequence that was not one of the given candidates approximately 14\% of the time. Most commonly, these hallucinated candidates were in the form of single noun phrases that were similar to segments of one or more of the given candidates.

The reason for this is likely a task mismatch. Decoder-only models such as ChatGPT are intended to generate sequences of text, which does not align well with the task of reranking.

\section{Conclusion}\label{sec:conclusion}
We have introduced a novel generate-and-rerank approach for generating high-quality natural language utterances from LFs using LLMs. Our approach is flexible and can be easily applied to diverse datasets and tasks. In addition, we have performed an analysis of the current popular evaluation metrics for NLG and selected the best metrics for the training and evaluation of our reranker. Our extensive experiments show that our reranker, which uses a loss function that compares individual candidates against one another, improves the quality of generated natural language in both fluency and semantic faithfulness in terms of the selected metrics on different evaluation datasets.

\section*{Limitations}
The results presented in Section \ref{sec:results} demonstrate that our reranker improves the quality of natural language text generated from LFs. However, the applicability of our method is somewhat limited by the choice of Codex as the generator model.

Firstly, Codex requires a lot of computation resources due to its size of 175 billion parameters \citep{Chen2021EvaluatingLL}, and a lot of time to generate candidates. A smaller model would be able to generate candidates much more efficiently, although those candidates would likely be lower quality. Further experimentation is required to determine whether the reranker's performance can make up for a weaker generator.

Secondly, it seems likely that the majority of the natural language data that appears in Codex's pre-training is in English, so our approach probably does not transfer well to other languages without modification. It may be beneficial to further explore this problem using a generator model with multilingual pre-training.

Another issue is that the reranker we introduce in this work, as we have formulated, may suffer from a lack of composition generalization abilities, as we note in Section \ref{sec:results}. The performance of a reranker in this setting may benefit from techniques used to improve compositional generalization in semantic parsers, such as the application of synthetic data \citep[][\textit{inter alia}]{wang-etal-2015-building, herzig-berant-2019-dont, yu2021grappa, wang-etal-2021-learning-synthesize, akyurek-andreas-2023-lexsym, li-etal-2023-learning} or the use of supervised attention \citep{yin-etal-2021-compositional}.

This approach could further be improved with the use of more reliable automated metrics. Our evaluation in Section \ref{sec:metric} found that the best performing combination of metrics had a top-1 accuracy of 81.4\% and a ranking accuracy of 84.22\%, which indicates that a fair number of the ranking decisions made by this combined metric were incorrect. However, due to time constraints, this study includes only one human annotator for our metric evaluation set, which hampers the reliability of our analysis of automatic metrics. Further exploration is needed to assess the alignment between different (combinations of) automatic metrics and human judgement of semantic consistency and fluency in this task. Additionally, there is much ongoing research in the creation and evaluation of automated metrics, and advances in this work would likely to translate to stronger performance of the method we have presented here.

\section*{Acknowledgements}
We express our deepest gratitude to David McGee, whose feedback has been critical in the development of this work. We are also grateful to the anonymous reviewers for their thoughtful comments.

\bibliography{anthology,custom}

\begin{thebibliography}{58}
\expandafter\ifx\csname natexlab\endcsname\relax\def\natexlab#1{#1}\fi

\bibitem[{Akyurek and Andreas(2023)}]{akyurek-andreas-2023-lexsym}
Ekin Akyurek and Jacob Andreas. 2023.
\newblock \href {https://doi.org/10.18653/v1/2023.acl-long.38} {{L}ex{S}ym:
  Compositionality as lexical symmetry}.
\newblock In \emph{Proceedings of the 61st Annual Meeting of the Association
  for Computational Linguistics (Volume 1: Long Papers)}, pages 639--657,
  Toronto, Canada. Association for Computational Linguistics.

\bibitem[{Amrhein et~al.(2022)Amrhein, Moghe, and Guillou}]{Amrhein2022ACESTA}
Chantal Amrhein, Nikita Moghe, and Liane Guillou. 2022.
\newblock Aces: Translation accuracy challenge sets for evaluating machine
  translation metrics.
\newblock \emph{ArXiv}, abs/2210.15615.

\bibitem[{Arcadinho et~al.(2022)Arcadinho, Apar{\'i}cio, Veiga, and
  Alegria}]{Arcadinho2022T5QLTL}
Samuel Arcadinho, David~Oliveira Apar{\'i}cio, Hugo Veiga, and Ant{\'o}nio
  Alegria. 2022.
\newblock T5ql: Taming language models for sql generation.
\newblock \emph{ArXiv}, abs/2209.10254.

\bibitem[{Bhattacharyya et~al.(2021)Bhattacharyya, Rooshenas, Naskar, Sun,
  Iyyer, and McCallum}]{bhattacharyya-etal-2021-energy}
Sumanta Bhattacharyya, Amirmohammad Rooshenas, Subhajit Naskar, Simeng Sun,
  Mohit Iyyer, and Andrew McCallum. 2021.
\newblock \href {https://doi.org/10.18653/v1/2021.acl-long.349} {Energy-based
  reranking: Improving neural machine translation using energy-based models}.
\newblock In \emph{Proceedings of the 59th Annual Meeting of the Association
  for Computational Linguistics and the 11th International Joint Conference on
  Natural Language Processing (Volume 1: Long Papers)}, pages 4528--4537,
  Online. Association for Computational Linguistics.

\bibitem[{Califf and Mooney(1999)}]{Califf1999RelationalLO}
Mary~Elaine Califf and Raymond~J. Mooney. 1999.
\newblock Relational learning of pattern-match rules for information
  extraction.
\newblock In \emph{Conference on Computational Natural Language Learning}.

\bibitem[{Chen et~al.(2021)Chen, Tworek, Jun, Yuan, Ponde, Kaplan, Edwards,
  Burda, Joseph, Brockman, Ray, Puri, Krueger, Petrov, Khlaaf, Sastry, Mishkin,
  Chan, Gray, Ryder, Pavlov, Power, Kaiser, Bavarian, Winter, Tillet, Such,
  Cummings, Plappert, Chantzis, Barnes, Herbert-Voss, Guss, Nichol, Babuschkin,
  Balaji, Jain, Carr, Leike, Achiam, Misra, Morikawa, Radford, Knight,
  Brundage, Murati, Mayer, Welinder, McGrew, Amodei, McCandlish, Sutskever, and
  Zaremba}]{Chen2021EvaluatingLL}
Mark Chen, Jerry Tworek, Heewoo Jun, Qiming Yuan, Henrique Ponde, Jared Kaplan,
  Harrison Edwards, Yura Burda, Nicholas Joseph, Greg Brockman, Alex Ray, Raul
  Puri, Gretchen Krueger, Michael Petrov, Heidy Khlaaf, Girish Sastry, Pamela
  Mishkin, Brooke Chan, Scott Gray, Nick Ryder, Mikhail Pavlov, Alethea Power,
  Lukasz Kaiser, Mohammad Bavarian, Clemens Winter, Philippe Tillet,
  Felipe~Petroski Such, David~W. Cummings, Matthias Plappert, Fotios Chantzis,
  Elizabeth Barnes, Ariel Herbert-Voss, William~H. Guss, Alex Nichol, Igor
  Babuschkin, S.~Arun Balaji, Shantanu Jain, Andrew Carr, Jan Leike, Joshua
  Achiam, Vedant Misra, Evan Morikawa, Alec Radford, Matthew~M. Knight, Miles
  Brundage, Mira Murati, Katie Mayer, Peter Welinder, Bob McGrew, Dario Amodei,
  Sam McCandlish, Ilya Sutskever, and Wojciech Zaremba. 2021.
\newblock Evaluating large language models trained on code.
\newblock \emph{ArXiv}, abs/2107.03374.

\bibitem[{Chen et~al.(2020{\natexlab{a}})Chen, Chen, Zha, Zhou, Zhang,
  Sundaresan, and Wang}]{chen-etal-2020-logic2text}
Zhiyu Chen, Wenhu Chen, Hanwen Zha, Xiyou Zhou, Yunkai Zhang, Sairam
  Sundaresan, and William~Yang Wang. 2020{\natexlab{a}}.
\newblock \href {https://doi.org/10.18653/v1/2020.findings-emnlp.190}
  {{L}ogic2{T}ext: High-fidelity natural language generation from logical
  forms}.
\newblock In \emph{Findings of the Association for Computational Linguistics:
  EMNLP 2020}, pages 2096--2111, Online. Association for Computational
  Linguistics.

\bibitem[{Chen et~al.(2020{\natexlab{b}})Chen, Eavani, Chen, Liu, and
  Wang}]{chen-etal-2020-shot}
Zhiyu Chen, Harini Eavani, Wenhu Chen, Yinyin Liu, and William~Yang Wang.
  2020{\natexlab{b}}.
\newblock \href {https://doi.org/10.18653/v1/2020.acl-main.18} {Few-shot {NLG}
  with pre-trained language model}.
\newblock In \emph{Proceedings of the 58th Annual Meeting of the Association
  for Computational Linguistics}, pages 183--190, Online. Association for
  Computational Linguistics.

\bibitem[{Cobbe et~al.(2021)Cobbe, Kosaraju, Bavarian, Chen, Jun, Kaiser,
  Plappert, Tworek, Hilton, Nakano, Hesse, and Schulman}]{cobbe2021training}
Karl Cobbe, Vineet Kosaraju, Mohammad Bavarian, Mark Chen, Heewoo Jun, Lukasz
  Kaiser, Matthias Plappert, Jerry Tworek, Jacob Hilton, Reiichiro Nakano,
  Christopher Hesse, and John Schulman. 2021.
\newblock \href {http://arxiv.org/abs/2110.14168} {Training verifiers to solve
  math word problems}.

\bibitem[{Deng et~al.(2020)Deng, Bakhtin, Ott, Szlam, and
  Ranzato}]{Deng2020Residual}
Yuntian Deng, Anton Bakhtin, Myle Ott, Arthur Szlam, and Marc'Aurelio Ranzato.
  2020.
\newblock \href {https://openreview.net/forum?id=B1l4SgHKDH} {Residual
  energy-based models for text generation}.
\newblock In \emph{International Conference on Learning Representations}.

\bibitem[{Drozdov et~al.(2022)Drozdov, Scharli, Akyuurek, Scales, Song, Chen,
  Bousquet, and Zhou}]{Drozdov2022CompositionalSP}
Andrew Drozdov, Nathanael Scharli, Ekin Akyuurek, Nathan Scales, Xinying Song,
  Xinyun Chen, Olivier Bousquet, and Denny Zhou. 2022.
\newblock Compositional semantic parsing with large language models.
\newblock \emph{ArXiv}, abs/2209.15003.

\bibitem[{Feng et~al.(2020)Feng, Guo, Tang, Duan, Feng, Gong, Shou, Qin, Liu,
  Jiang, and Zhou}]{feng-etal-2020-codebert}
Zhangyin Feng, Daya Guo, Duyu Tang, Nan Duan, Xiaocheng Feng, Ming Gong, Linjun
  Shou, Bing Qin, Ting Liu, Daxin Jiang, and Ming Zhou. 2020.
\newblock \href {https://doi.org/10.18653/v1/2020.findings-emnlp.139}
  {{C}ode{BERT}: A pre-trained model for programming and natural languages}.
\newblock In \emph{Findings of the Association for Computational Linguistics:
  EMNLP 2020}, pages 1536--1547, Online. Association for Computational
  Linguistics.

\bibitem[{Finegan-Dollak et~al.(2018)Finegan-Dollak, Kummerfeld, Zhang,
  Ramanathan, Sadasivam, Zhang, and Radev}]{finegan2018improving}
Catherine Finegan-Dollak, Jonathan~K Kummerfeld, Li~Zhang, Karthik Ramanathan,
  Sesh Sadasivam, Rui Zhang, and Dragomir Radev. 2018.
\newblock Improving text-to-sql evaluation methodology.
\newblock In \emph{Proceedings of the 56th Annual Meeting of the Association
  for Computational Linguistics (Volume 1: Long Papers)}, pages 351--360.

\bibitem[{Freitag et~al.(2022)Freitag, Rei, Mathur, kiu Lo, Stewart, Avramidis,
  Kocmi, Foster, Lavie, and Martins}]{WMT22}
Markus Freitag, Ricardo Rei, Nitika Mathur, Chi kiu Lo, Craig Stewart,
  Eleftherios Avramidis, Tom Kocmi, George Foster, Alon Lavie, and André
  Martins. 2022.
\newblock \href {https://www.statmt.org/wmt22/pdf/2022.wmt-1.2.pdf} {Results of
  wmt22 metrics shared task: Stop using bleu - neural metrics are better and
  more robust}.
\newblock In \emph{Proceedings of the Seventh Conference on Machine
  Translation}, pages 46--68, Abu Dhabi.

\bibitem[{Gardent et~al.(2017)Gardent, Shimorina, Narayan, and
  Perez-Beltrachini}]{gardent-etal-2017-webnlg}
Claire Gardent, Anastasia Shimorina, Shashi Narayan, and Laura
  Perez-Beltrachini. 2017.
\newblock \href {https://doi.org/10.18653/v1/W17-3518} {The {W}eb{NLG}
  challenge: Generating text from {RDF} data}.
\newblock In \emph{Proceedings of the 10th International Conference on Natural
  Language Generation}, pages 124--133, Santiago de Compostela, Spain.
  Association for Computational Linguistics.

\bibitem[{Gehrmann et~al.(2021)Gehrmann, Adewumi, Aggarwal, Ammanamanchi,
  Aremu, Bosselut, Chandu, Clinciu, Das, Dhole, Du, Durmus, Du{\v{s}}ek,
  Emezue, Gangal, Garbacea, Hashimoto, Hou, Jernite, Jhamtani, Ji, Jolly, Kale,
  Kumar, Ladhak, Madaan, Maddela, Mahajan, Mahamood, Majumder, Martins,
  McMillan-Major, Mille, van Miltenburg, Nadeem, Narayan, Nikolaev,
  Niyongabo~Rubungo, Osei, Parikh, Perez-Beltrachini, Rao, Raunak, Rodriguez,
  Santhanam, Sedoc, Sellam, Shaikh, Shimorina, Sobrevilla~Cabezudo, Strobelt,
  Subramani, Xu, Yang, Yerukola, and Zhou}]{gehrmann-etal-2021-gem}
Sebastian Gehrmann, Tosin Adewumi, Karmanya Aggarwal, Pawan~Sasanka
  Ammanamanchi, Anuoluwapo Aremu, Antoine Bosselut, Khyathi~Raghavi Chandu,
  Miruna-Adriana Clinciu, Dipanjan Das, Kaustubh Dhole, Wanyu Du, Esin Durmus,
  Ond{\v{r}}ej Du{\v{s}}ek, Chris~Chinenye Emezue, Varun Gangal, Cristina
  Garbacea, Tatsunori Hashimoto, Yufang Hou, Yacine Jernite, Harsh Jhamtani,
  Yangfeng Ji, Shailza Jolly, Mihir Kale, Dhruv Kumar, Faisal Ladhak, Aman
  Madaan, Mounica Maddela, Khyati Mahajan, Saad Mahamood, Bodhisattwa~Prasad
  Majumder, Pedro~Henrique Martins, Angelina McMillan-Major, Simon Mille, Emiel
  van Miltenburg, Moin Nadeem, Shashi Narayan, Vitaly Nikolaev, Andre
  Niyongabo~Rubungo, Salomey Osei, Ankur Parikh, Laura Perez-Beltrachini,
  Niranjan~Ramesh Rao, Vikas Raunak, Juan~Diego Rodriguez, Sashank Santhanam,
  Jo{\~a}o Sedoc, Thibault Sellam, Samira Shaikh, Anastasia Shimorina,
  Marco~Antonio Sobrevilla~Cabezudo, Hendrik Strobelt, Nishant Subramani, Wei
  Xu, Diyi Yang, Akhila Yerukola, and Jiawei Zhou. 2021.
\newblock \href {https://doi.org/10.18653/v1/2021.gem-1.10} {The {GEM}
  benchmark: Natural language generation, its evaluation and metrics}.
\newblock In \emph{Proceedings of the 1st Workshop on Natural Language
  Generation, Evaluation, and Metrics (GEM 2021)}, pages 96--120, Online.
  Association for Computational Linguistics.

\bibitem[{Harkous et~al.(2020)Harkous, Groves, and
  Saffari}]{harkous-etal-2020-text}
Hamza Harkous, Isabel Groves, and Amir Saffari. 2020.
\newblock \href {https://doi.org/10.18653/v1/2020.coling-main.218} {Have your
  text and use it too! end-to-end neural data-to-text generation with semantic
  fidelity}.
\newblock In \emph{Proceedings of the 28th International Conference on
  Computational Linguistics}, pages 2410--2424, Barcelona, Spain (Online).
  International Committee on Computational Linguistics.

\bibitem[{Heidari et~al.(2021)Heidari, Einolghozati, Jain, Batra, Callender,
  Arun, Mei, Gupta, Donmez, Bhardwaj, Kumar, and
  White}]{heidari-etal-2021-getting}
Peyman Heidari, Arash Einolghozati, Shashank Jain, Soumya Batra, Lee Callender,
  Ankit Arun, Shawn Mei, Sonal Gupta, Pinar Donmez, Vikas Bhardwaj, Anuj Kumar,
  and Michael White. 2021.
\newblock \href {https://aclanthology.org/2021.sigdial-1.8} {Getting to
  production with few-shot natural language generation models}.
\newblock In \emph{Proceedings of the 22nd Annual Meeting of the Special
  Interest Group on Discourse and Dialogue}, pages 66--76, Singapore and
  Online. Association for Computational Linguistics.

\bibitem[{Herzig and Berant(2019)}]{herzig-berant-2019-dont}
Jonathan Herzig and Jonathan Berant. 2019.
\newblock \href {https://doi.org/10.18653/v1/D19-1394} {Don{'}t paraphrase,
  detect! rapid and effective data collection for semantic parsing}.
\newblock In \emph{Proceedings of the 2019 Conference on Empirical Methods in
  Natural Language Processing and the 9th International Joint Conference on
  Natural Language Processing (EMNLP-IJCNLP)}, pages 3810--3820, Hong Kong,
  China. Association for Computational Linguistics.

\bibitem[{Keysers et~al.(2019)Keysers, Sch{\"a}rli, Scales, Buisman, Furrer,
  Kashubin, Momchev, Sinopalnikov, Stafiniak, Tihon, Tsarkov, Wang, van Zee,
  and Bousquet}]{Keysers2019MeasuringCG}
Daniel Keysers, Nathanael Sch{\"a}rli, Nathan Scales, Hylke Buisman, Daniel
  Furrer, Sergii Kashubin, Nikola Momchev, Danila Sinopalnikov, Lukasz
  Stafiniak, Tibor Tihon, Dmitry Tsarkov, Xiao Wang, Marc van Zee, and Olivier
  Bousquet. 2019.
\newblock Measuring compositional generalization: A comprehensive method on
  realistic data.
\newblock \emph{ArXiv}, abs/1912.09713.

\bibitem[{Kingma and Ba(2014)}]{Kingma2014AdamAM}
Diederik~P. Kingma and Jimmy Ba. 2014.
\newblock Adam: A method for stochastic optimization.
\newblock \emph{CoRR}, abs/1412.6980.

\bibitem[{Kullback and Leibler(1951)}]{kullback1951information}
Solomon Kullback and Richard~A Leibler. 1951.
\newblock On information and sufficiency.
\newblock \emph{The annals of mathematical statistics}, 22(1):79--86.

\bibitem[{Langkilde-Geary(2002)}]{langkilde-geary-2002-empirical}
Irene Langkilde-Geary. 2002.
\newblock \href {https://aclanthology.org/W02-2103} {An empirical verification
  of coverage and correctness for a general-purpose sentence generator}.
\newblock In \emph{Proceedings of the International Natural Language Generation
  Conference}, pages 17--24, Harriman, New York, USA. Association for
  Computational Linguistics.

\bibitem[{Lee et~al.(2021)Lee, Auli, and
  Ranzato}]{lee-etal-2021-discriminative}
Ann Lee, Michael Auli, and Marc{'}Aurelio Ranzato. 2021.
\newblock \href {https://doi.org/10.18653/v1/2021.acl-long.563} {Discriminative
  reranking for neural machine translation}.
\newblock In \emph{Proceedings of the 59th Annual Meeting of the Association
  for Computational Linguistics and the 11th International Joint Conference on
  Natural Language Processing (Volume 1: Long Papers)}, pages 7250--7264,
  Online. Association for Computational Linguistics.

\bibitem[{Li et~al.(2023)Li, Wei, and Lian}]{li-etal-2023-learning}
Zhaoyi Li, Ying Wei, and Defu Lian. 2023.
\newblock \href {https://doi.org/10.18653/v1/2023.acl-long.157} {Learning to
  substitute spans towards improving compositional generalization}.
\newblock In \emph{Proceedings of the 61st Annual Meeting of the Association
  for Computational Linguistics (Volume 1: Long Papers)}, pages 2791--2811,
  Toronto, Canada. Association for Computational Linguistics.

\bibitem[{Li et~al.(2022)Li, Qu, Xu, Wu, Zhan, and Haffari}]{li2022variational}
Zhuang Li, Lizhen Qu, Qiongkai Xu, Tongtong Wu, Tianyang Zhan, and Gholamreza
  Haffari. 2022.
\newblock Variational autoencoder with disentanglement priors for low-resource
  task-specific natural language generation.
\newblock In \emph{Proceedings of the 2022 Conference on Empirical Methods in
  Natural Language Processing}, pages 10335--10356.

\bibitem[{Liu and Liu(2021)}]{liu-liu-2021-simcls}
Yixin Liu and Pengfei Liu. 2021.
\newblock \href {https://doi.org/10.18653/v1/2021.acl-short.135} {{S}im{CLS}: A
  simple framework for contrastive learning of abstractive summarization}.
\newblock In \emph{Proceedings of the 59th Annual Meeting of the Association
  for Computational Linguistics and the 11th International Joint Conference on
  Natural Language Processing (Volume 2: Short Papers)}, pages 1065--1072,
  Online. Association for Computational Linguistics.

\bibitem[{Ma et~al.(2021)Ma, Chen, Cao, Chen, Chen, and
  Yu}]{Ma2021RelationAwareGT}
Da~Ma, Xingyu Chen, Ruisheng Cao, Zhi Chen, Lu~Chen, and Kai Yu. 2021.
\newblock Relation-aware graph transformer for sql-to-text generation.
\newblock \emph{Applied Sciences}.

\bibitem[{Moghe et~al.(2022)Moghe, Sherborne, Steedman, and
  Birch}]{Moghe2022ExtrinsicEO}
Nikita Moghe, Tom Sherborne, Mark Steedman, and Alexandra Birch. 2022.
\newblock Extrinsic evaluation of machine translation metrics.
\newblock \emph{ArXiv}, abs/2212.10297.

\bibitem[{Ngomo et~al.(2013)Ngomo, B{\"u}hmann, Unger, Lehmann, and
  Gerber}]{Ngomo2013SorryID}
Axel-Cyrille~Ngonga Ngomo, Lorenz B{\"u}hmann, Christina Unger, Jens Lehmann,
  and Daniel Gerber. 2013.
\newblock Sorry, i don't speak sparql: translating sparql queries into natural
  language.
\newblock \emph{Proceedings of the 22nd international conference on World Wide
  Web}.

\bibitem[{Ouyang et~al.(2022)Ouyang, Wu, Jiang, Almeida, Wainwright, Mishkin,
  Zhang, Agarwal, Slama, Ray, Schulman, Hilton, Kelton, Miller, Simens, Askell,
  Welinder, Christiano, Leike, and Lowe}]{Ouyang2022TrainingLM}
Long Ouyang, Jeff Wu, Xu~Jiang, Diogo Almeida, Carroll~L. Wainwright, Pamela
  Mishkin, Chong Zhang, Sandhini Agarwal, Katarina Slama, Alex Ray, John
  Schulman, Jacob Hilton, Fraser Kelton, Luke~E. Miller, Maddie Simens, Amanda
  Askell, Peter Welinder, Paul~Francis Christiano, Jan Leike, and Ryan~J. Lowe.
  2022.
\newblock Training language models to follow instructions with human feedback.
\newblock \emph{ArXiv}, abs/2203.02155.

\bibitem[{Papineni et~al.(2002)Papineni, Roukos, Ward, and
  Zhu}]{papineni-etal-2002-bleu}
Kishore Papineni, Salim Roukos, Todd Ward, and Wei-Jing Zhu. 2002.
\newblock \href {https://doi.org/10.3115/1073083.1073135} {{B}leu: a method for
  automatic evaluation of machine translation}.
\newblock In \emph{Proceedings of the 40th Annual Meeting of the Association
  for Computational Linguistics}, pages 311--318, Philadelphia, Pennsylvania,
  USA. Association for Computational Linguistics.

\bibitem[{Parikh et~al.(2020)Parikh, Wang, Gehrmann, Faruqui, Dhingra, Yang,
  and Das}]{parikh-etal-2020-totto}
Ankur Parikh, Xuezhi Wang, Sebastian Gehrmann, Manaal Faruqui, Bhuwan Dhingra,
  Diyi Yang, and Dipanjan Das. 2020.
\newblock \href {https://doi.org/10.18653/v1/2020.emnlp-main.89} {{ToTTo}: A
  controlled table-to-text generation dataset}.
\newblock In \emph{Proceedings of the 2020 Conference on Empirical Methods in
  Natural Language Processing (EMNLP)}, pages 1173--1186, Online. Association
  for Computational Linguistics.

\bibitem[{Pu et~al.(2021)Pu, Chung, Parikh, Gehrmann, and
  Sellam}]{pu-etal-2021-learning}
Amy Pu, Hyung~Won Chung, Ankur Parikh, Sebastian Gehrmann, and Thibault Sellam.
  2021.
\newblock \href {https://doi.org/10.18653/v1/2021.emnlp-main.58} {Learning
  compact metrics for {MT}}.
\newblock In \emph{Proceedings of the 2021 Conference on Empirical Methods in
  Natural Language Processing}, pages 751--762, Online and Punta Cana,
  Dominican Republic. Association for Computational Linguistics.

\bibitem[{Ribeiro et~al.(2019)Ribeiro, Gardent, and
  Gurevych}]{Ribeiro2019EnhancingAG}
Leonardo F.~R. Ribeiro, Claire Gardent, and Iryna Gurevych. 2019.
\newblock Enhancing amr-to-text generation with dual graph representations.
\newblock In \emph{Conference on Empirical Methods in Natural Language
  Processing}.

\bibitem[{Ribeiro et~al.(2021{\natexlab{a}})Ribeiro, Zhang, and
  Gurevych}]{ribeiro-etal-2021-structural}
Leonardo F.~R. Ribeiro, Yue Zhang, and Iryna Gurevych. 2021{\natexlab{a}}.
\newblock \href {https://doi.org/10.18653/v1/2021.emnlp-main.351} {Structural
  adapters in pretrained language models for {AMR}-to-{T}ext generation}.
\newblock In \emph{Proceedings of the 2021 Conference on Empirical Methods in
  Natural Language Processing}, pages 4269--4282, Online and Punta Cana,
  Dominican Republic. Association for Computational Linguistics.

\bibitem[{Ribeiro et~al.(2021{\natexlab{b}})Ribeiro, Schmitt, Sch{\"u}tze, and
  Gurevych}]{ribeiro2021investigating}
Leonardo~FR Ribeiro, Martin Schmitt, Hinrich Sch{\"u}tze, and Iryna Gurevych.
  2021{\natexlab{b}}.
\newblock Investigating pretrained language models for graph-to-text
  generation.
\newblock In \emph{Proceedings of the 3rd Workshop on Natural Language
  Processing for Conversational AI}, pages 211--227.

\bibitem[{Sellam et~al.(2020)Sellam, Das, and Parikh}]{sellam-etal-2020-bleurt}
Thibault Sellam, Dipanjan Das, and Ankur Parikh. 2020.
\newblock \href {https://doi.org/10.18653/v1/2020.acl-main.704} {{BLEURT}:
  Learning robust metrics for text generation}.
\newblock In \emph{Proceedings of the 58th Annual Meeting of the Association
  for Computational Linguistics}, pages 7881--7892, Online. Association for
  Computational Linguistics.

\bibitem[{Shin et~al.(2021)Shin, Lin, Thomson, Chen, Roy, Platanios, Pauls,
  Klein, Eisner, and Van~Durme}]{shin-etal-2021-constrained}
Richard Shin, Christopher Lin, Sam Thomson, Charles Chen, Subhro Roy,
  Emmanouil~Antonios Platanios, Adam Pauls, Dan Klein, Jason Eisner, and
  Benjamin Van~Durme. 2021.
\newblock \href {https://doi.org/10.18653/v1/2021.emnlp-main.608} {Constrained
  language models yield few-shot semantic parsers}.
\newblock In \emph{Proceedings of the 2021 Conference on Empirical Methods in
  Natural Language Processing}, pages 7699--7715, Online and Punta Cana,
  Dominican Republic. Association for Computational Linguistics.

\bibitem[{Shin and Van~Durme(2022)}]{shin-van-durme-2022-shot}
Richard Shin and Benjamin Van~Durme. 2022.
\newblock \href {https://doi.org/10.18653/v1/2022.naacl-main.396} {Few-shot
  semantic parsing with language models trained on code}.
\newblock In \emph{Proceedings of the 2022 Conference of the North American
  Chapter of the Association for Computational Linguistics: Human Language
  Technologies}, pages 5417--5425, Seattle, United States. Association for
  Computational Linguistics.

\bibitem[{Shiri et~al.(2022)Shiri, Zhuo, Li, Pan, Wang, Haffari, Li, and
  Nguyen}]{shiri2022paraphrasing}
Fatemeh Shiri, Terry~Yue Zhuo, Zhuang Li, Shirui Pan, Weiqing Wang, Reza
  Haffari, Yuan-Fang Li, and Van Nguyen. 2022.
\newblock Paraphrasing techniques for maritime qa system.
\newblock In \emph{2022 25th International Conference on Information Fusion
  (FUSION)}, pages 1--8. IEEE.

\bibitem[{Song et~al.(2018)Song, Zhang, Wang, and
  Gildea}]{song-etal-2018-graph}
Linfeng Song, Yue Zhang, Zhiguo Wang, and Daniel Gildea. 2018.
\newblock \href {https://doi.org/10.18653/v1/P18-1150} {A graph-to-sequence
  model for {AMR}-to-text generation}.
\newblock In \emph{Proceedings of the 56th Annual Meeting of the Association
  for Computational Linguistics (Volume 1: Long Papers)}, pages 1616--1626,
  Melbourne, Australia. Association for Computational Linguistics.

\bibitem[{Suzgun et~al.(2022)Suzgun, Melas-Kyriazi, and
  Jurafsky}]{suzgun-etal-2022-prompt}
Mirac Suzgun, Luke Melas-Kyriazi, and Dan Jurafsky. 2022.
\newblock \href {https://aclanthology.org/2022.emnlp-main.141}
  {Prompt-and-rerank: A method for zero-shot and few-shot arbitrary textual
  style transfer with small language models}.
\newblock In \emph{Proceedings of the 2022 Conference on Empirical Methods in
  Natural Language Processing}, pages 2195--2222, Abu Dhabi, United Arab
  Emirates. Association for Computational Linguistics.

\bibitem[{Thompson and Post(2020)}]{thompson-post-2020-automatic}
Brian Thompson and Matt Post. 2020.
\newblock \href {https://doi.org/10.18653/v1/2020.emnlp-main.8} {Automatic
  machine translation evaluation in many languages via zero-shot paraphrasing}.
\newblock In \emph{Proceedings of the 2020 Conference on Empirical Methods in
  Natural Language Processing (EMNLP)}, pages 90--121, Online. Association for
  Computational Linguistics.

\bibitem[{Wang et~al.(2021{\natexlab{a}})Wang, Yin, Lin, and
  Xiong}]{wang2021learning}
Bailin Wang, Wenpeng Yin, Xi~Victoria Lin, and Caiming Xiong.
  2021{\natexlab{a}}.
\newblock Learning to synthesize data for semantic parsing.
\newblock In \emph{Proceedings of the 2021 Conference of the North American
  Chapter of the Association for Computational Linguistics: Human Language
  Technologies}, pages 2760--2766.

\bibitem[{Wang et~al.(2021{\natexlab{b}})Wang, Yin, Lin, and
  Xiong}]{wang-etal-2021-learning-synthesize}
Bailin Wang, Wenpeng Yin, Xi~Victoria Lin, and Caiming Xiong.
  2021{\natexlab{b}}.
\newblock \href {https://doi.org/10.18653/v1/2021.naacl-main.220} {Learning to
  synthesize data for semantic parsing}.
\newblock In \emph{Proceedings of the 2021 Conference of the North American
  Chapter of the Association for Computational Linguistics: Human Language
  Technologies}, pages 2760--2766, Online. Association for Computational
  Linguistics.

\bibitem[{Wang et~al.(2022)Wang, Wei, Schuurmans, Le, hsin Chi, and
  Zhou}]{Wang2022SelfConsistencyIC}
Xuezhi Wang, Jason Wei, Dale Schuurmans, Quoc Le, Ed~Huai hsin Chi, and Denny
  Zhou. 2022.
\newblock Self-consistency improves chain of thought reasoning in language
  models.
\newblock \emph{ArXiv}, abs/2203.11171.

\bibitem[{Wang et~al.(2021{\natexlab{c}})Wang, Wang, Joty, and
  Hoi}]{wang-etal-2021-codet5}
Yue Wang, Weishi Wang, Shafiq Joty, and Steven~C.H. Hoi. 2021{\natexlab{c}}.
\newblock \href {https://doi.org/10.18653/v1/2021.emnlp-main.685} {{C}ode{T}5:
  Identifier-aware unified pre-trained encoder-decoder models for code
  understanding and generation}.
\newblock In \emph{Proceedings of the 2021 Conference on Empirical Methods in
  Natural Language Processing}, pages 8696--8708, Online and Punta Cana,
  Dominican Republic. Association for Computational Linguistics.

\bibitem[{Wang et~al.(2015)Wang, Berant, and Liang}]{wang-etal-2015-building}
Yushi Wang, Jonathan Berant, and Percy Liang. 2015.
\newblock \href {https://doi.org/10.3115/v1/P15-1129} {Building a semantic
  parser overnight}.
\newblock In \emph{Proceedings of the 53rd Annual Meeting of the Association
  for Computational Linguistics and the 7th International Joint Conference on
  Natural Language Processing (Volume 1: Long Papers)}, pages 1332--1342,
  Beijing, China. Association for Computational Linguistics.

\bibitem[{Xu et~al.(2018)Xu, Wu, Wang, Feng, and Sheinin}]{xu-etal-2018-sql}
Kun Xu, Lingfei Wu, Zhiguo Wang, Yansong Feng, and Vadim Sheinin. 2018.
\newblock \href {https://doi.org/10.18653/v1/D18-1112} {{SQL}-to-text
  generation with graph-to-sequence model}.
\newblock In \emph{Proceedings of the 2018 Conference on Empirical Methods in
  Natural Language Processing}, pages 931--936, Brussels, Belgium. Association
  for Computational Linguistics.

\bibitem[{Yin et~al.(2021)Yin, Fang, Neubig, Pauls, Platanios, Su, Thomson, and
  Andreas}]{yin-etal-2021-compositional}
Pengcheng Yin, Hao Fang, Graham Neubig, Adam Pauls, Emmanouil~Antonios
  Platanios, Yu~Su, Sam Thomson, and Jacob Andreas. 2021.
\newblock \href {https://doi.org/10.18653/v1/2021.naacl-main.225}
  {Compositional generalization for neural semantic parsing via span-level
  supervised attention}.
\newblock In \emph{Proceedings of the 2021 Conference of the North American
  Chapter of the Association for Computational Linguistics: Human Language
  Technologies}, pages 2810--2823, Online. Association for Computational
  Linguistics.

\bibitem[{Yu et~al.(2021)Yu, Wu, Lin, Wang, Tan, Yang, Radev, Socher, and
  Xiong}]{yu2021grappa}
Tao Yu, Chien-Sheng Wu, Xi~Victoria Lin, Bailin Wang, Yi~Chern Tan, Xinyi Yang,
  Dragomir Radev, Richard Socher, and Caiming Xiong. 2021.
\newblock \href {http://arxiv.org/abs/2009.13845} {Grappa: Grammar-augmented
  pre-training for table semantic parsing}.

\bibitem[{Yu et~al.(2019)Yu, Zhang, Er, Li, Xue, Pang, Lin, Tan, Shi, Li
  et~al.}]{yu2019cosql}
Tao Yu, Rui Zhang, He~Yang Er, Suyi Li, Eric Xue, Bo~Pang, Xi~Victoria Lin,
  Yi~Chern Tan, Tianze Shi, Zihan Li, et~al. 2019.
\newblock Cosql: A conversational text-to-sql challenge towards cross-domain
  natural language interfaces to databases.
\newblock \emph{arXiv preprint arXiv:1909.05378}.

\bibitem[{Yuan et~al.(2021)Yuan, Neubig, and Liu}]{Yuan2021BARTScoreEG}
Weizhe Yuan, Graham Neubig, and Pengfei Liu. 2021.
\newblock Bartscore: Evaluating generated text as text generation.
\newblock \emph{ArXiv}, abs/2106.11520.

\bibitem[{Zelle and Mooney(1996)}]{Zelle1996LearningTP}
John~M. Zelle and Raymond~J. Mooney. 1996.
\newblock Learning to parse database queries using inductive logic programming.
\newblock In \emph{AAAI/IAAI, Vol. 2}.

\bibitem[{Zhang et~al.(2019)Zhang, Kishore, Wu, Weinberger, and
  Artzi}]{Zhang2019BERTScoreET}
Tianyi Zhang, Varsha Kishore, Felix Wu, Kilian~Q. Weinberger, and Yoav Artzi.
  2019.
\newblock Bertscore: Evaluating text generation with bert.
\newblock \emph{ArXiv}, abs/1904.09675.

\bibitem[{Zhu et~al.(2019)Zhu, Li, Zhu, Qian, Zhang, and
  Zhou}]{Zhu2019ModelingGS}
Jiehan Zhu, Junhui Li, Muhua Zhu, Longhua Qian, Min Zhang, and Guodong Zhou.
  2019.
\newblock Modeling graph structure in transformer for better amr-to-text
  generation.
\newblock In \emph{Conference on Empirical Methods in Natural Language
  Processing}.

\bibitem[{Zhuo et~al.(2023)Zhuo, Li, Huang, Shiri, Wang, Haffari, and
  Li}]{zhuo2023robustness}
Terry~Yue Zhuo, Zhuang Li, Yujin Huang, Fatemeh Shiri, Weiqing Wang, Gholamreza
  Haffari, and Yuan-Fang Li. 2023.
\newblock On robustness of prompt-based semantic parsing with large pre-trained
  language model: An empirical study on codex.
\newblock In \emph{Proceedings of the 17th Conference of the European Chapter
  of the Association for Computational Linguistics}, pages 1090--1102.

\end{thebibliography}
\bibliographystyle{acl_natbib}

\newpage

\appendix

\section{Reranker Training Details}
\label{sec:details}

The reranker model is trained for a maximum of 100 epochs with early stopping if the loss on the development set does not decrease after 10 epochs. Each batch comprises all of the candidates corresponding to a single logical form, so the batch size is equal to the size of the candidate list. We utilized Adam \citep{Kingma2014AdamAM} to optimize the model, with a learning rate of \num{1e-4}. The best model is determined as the one that produces the smallest loss on a held-out development set.

Hyperparameter tuning was conducted to determine the learning rate and the optimal epoch count for unfreezing the base model's final layer. The learning rates explored during this process were [\num{1e-5}, \num{5e-5}, \num{1e-4}, \num{5e-4}, \num{1e-3}], while the numbers of epochs before unfreezing considered were [1, 5, 10, 20]. The model's training was conducted on a single NVIDIA Tesla V100 GPU. The duration of training varied significantly depending on the size of the training dataset, ranging from a minimum of approximately 20 minutes to a maximum of around 35 hours.

\section{Generation from ChatGPT}
\label{sec:chatgpt}
To account for the fact that ChatGPT is optimized for chat functionality while Codex is not, we modify our generation prompt slightly. We use the same number of in-context examples (15) for both generators, but for ChatGPT we incorporate more natural language instruction to contextualize the examples and specifically prompt the model to generate eight unique candidates. The full prompt can be found in Appendix \ref{sec:chatgpt_gen_prompt}. To complete the task of reranking using in-context learning, we use a prompt that provides exemplars from the training set in order to condition the model on correct pairings of LFs and natural language. We present the prompt used for the reranking task in Appendix \ref{sec:chatgpt_rerank_prompt}.

\section{Sample Prompts}
\label{sec:sample_prompt}

\subsection{Codex generation prompt}\label{sec:codex_prompt}
Below is an example that illustrates the format of our prompts to Codex.

\begin{spverbatim}
    
# geo_query Dataset:

Query: answer ( longest ( intersection ( river , traverse_2 ( intersection ( state , next_to_2 ( m0 ) ) ) ) ) )
Question: what is the longest river that flows through a state that borders m0

Query: answer ( intersection ( state , next_to_2 ( largest_one ( population_1 , state ) ) ) )
Question: what are the states that border the state with the greatest population

Query: answer ( intersection ( river , traverse_2 ( m0 ) ) )
Question: what rivers run through m0

Query: answer ( count ( intersection ( state , low_point_2 ( lower_2 ( low_point_1 ( m0 ) ) ) ) ) )
Question: count the states which have elevations lower than what m0 has

Query: answer ( highest ( intersection ( place , loc_2 ( smallest_one ( population_1 , state ) ) ) ) )
Question: what is the highest point in the state with the smallest population

Query: answer ( intersection ( state , next_to_2 ( m0 ) ) )
Question: which states border m0

Query: answer ( density_1 ( intersection ( state , traverse_1 ( longest ( intersection ( river , loc_2 ( m0 ) ) ) ) ) ) )
Question: which is the density of the state that the largest river in the m0 runs through

Query: answer ( elevation_1 ( highest ( intersection ( place , loc_2 ( state ) ) ) ) )
Question: how high are the highest points of all the states

Query: answer ( count ( intersection ( state , loc_2 ( m0 ) ) ) )
Question: how many states are in the m0

Query: answer ( loc_1 ( m0 ) )
Question: where is m0

Query: answer ( intersection ( state , next_to_2 ( m0 ) ) )
Question: what state borders m0

Query: answer ( intersection ( state , loc_1 ( highest ( place ) ) ) )
Question: which state has the highest elevation

Query: answer ( intersection ( state , capital_2 ( m0 ) ) )
Question: what states capital is m0

Query: answer ( intersection ( state , next_to_2 ( m0 ) ) )
Question: what states surround m0

Query: answer ( count ( intersection ( river , loc_2 ( m0 ) ) ) )
Question: how many rivers are found in m0

Query: answer ( largest ( intersection ( state , loc_2 ( m0 ) ) ) )
Question:
\end{spverbatim}

\subsection{ChatGPT generation prompt}\label{sec:chatgpt_gen_prompt}

Below is an example that illustrates the format of our prompts to ChatGPT for generating natural language candidates. Most of the exemplars are elided here for brevity. This prompt uses the same number of exemplars as the prompt in Appendix \ref{sec:codex_prompt}, using the slightly modified form shown below.

\begin{spverbatim}
Here are some examples of query/question pairs from the GeoQuery data set.

logical form: answer ( longest ( intersection ( river , traverse_2 ( intersection ( state , next_to_2 ( m0 ) ) ) ) ) )
natural language: what is the longest river that flows through a state that borders m0

[...]

Please generate 8 natural language candidates for following logical form. Present your answer as a numbered list.    
logical form: answer ( largest ( intersection ( state , loc_2 ( m0 ) ) ) )
\end{spverbatim}

\subsection{ChatGPT reranking prompt}\label{sec:chatgpt_rerank_prompt}

Below is an example that illustrates the format of our prompts to ChatGPT for reranking natural language candidates. Most of the exemplars are elided here for brevity.

\begin{spverbatim}
Here are some examples of query/question pairs from the GeoQuery data set.

logical form: answer ( longest ( intersection ( river , traverse_2 ( intersection ( state , next_to_2 ( m0 ) ) ) ) ) )
natural language: what is the longest river that flows through a state that borders m0

[...]

I would like for you to rank some natural language candidates for the following logical form.    
logical form: answer ( largest ( intersection ( state , loc_2 ( m0 ) ) ) )

Here are the candidates:

Which state in m0 has the largest area?
What is the largest state that lies within m0?
Which state in m0 has the largest population?
What is the largest state found in m0 by area?
Which state in m0 is the largest in terms of land area?
What state located in m0 has the largest landmass?
What is the largest state located in m0 by size?
Which state in m0 has the highest number of inhabitants?

Which of these candidates is the best? Please return the text of the best candidate in quotation marks.
\end{spverbatim}

\section{Freebase Identifier Mapping}
\label{sec:mapping}
Table \ref{tab:mapping} shows the mapping we use to shorten Freebase IDs into strings that are easier to interpret. Most of this mapping originates in \citet{Drozdov2022CompositionalSP}.

\begin{table*}[ht]
    \small
    \centering
    \begin{tabular}{ll}
        \textbf{Freebase Identifier} & \textbf{Mapped String} \\
        \hline
ns:organization.organization.companies\_acquired/ns:business. & acquired \\
ns:organization.organization.acquired\_by/ns:business.acquisi & acquired\_by \\
ns:film.actor.film/ns:film.performance.film & starred\_in \\
ns:film.film\_art\_director.films\_art\_directed & art\_directed \\
ns:film.film.film\_art\_direction\_by & art\_direction\_by \\
ns:film.film.cinematography & cinematography\_by \\
ns:film.film\_costumer\_designer.costume\_design\_for\_film & costume\_designed \\
ns:film.film.costume\_design\_by & costume\_designed\_by \\
ns:film.director.film & directed \\
ns:film.film.directed\_by & directed\_by \\
ns:film.film.distributors/ns:film.film\_film\_distributor\_rela & distributed\_by \\
ns:film.film\_distributor.films\_distributed/ns:film.film\_film & distributed \\
ns:film.editor.film & edited \\
ns:film.film.edited\_by & edited\_by \\
ns:business.employer.employees/ns:business.employment\_tenure & employed \\
ns:people.person.employment\_history/ns:business.employment\_t & employed\_by \\
ns:film.producer.films\_executive\_produced & executive\_produced \\
ns:film.film.executive\_produced\_by & executive\_produced\_by \\
ns:organization.organization\_founder.organizations\_founded & founded \\
ns:organization.organization.founders & founded\_by \\
ns:people.person.gender & gender\_is \\
\^ns:people.person.gender & same\_gender\_as \\
ns:film.actor.film/ns:film.performance.character & portrayed \\
ns:people.person.nationality & nationality\_is \\
ns:film.film.prequel & sequel\_of \\
ns:film.film.sequel & prequel\_of \\
ns:influence.influence\_node.influenced & influenced \\
ns:influence.influence\_node.influenced\_by & influenced\_by \\
ns:people.person.spouse\_s/ns:people.marriage.spouse|ns:ficti & married\_to \\
\^ns:people.person.nationality & same\_nationality\_as \\
ns:people.person.children|ns:fictional\_universe.fictional\_ch & parent\_of \\
ns:people.person.parents|ns:fictional\_universe.fictional\_cha & child\_of \\
ns:film.producer.film|ns:film.production\_company.films & produced \\
ns:film.film.produced\_by|ns:film.film.production\_companies & produced\_by \\
ns:people.person.sibling\_s/ns:people.sibling\_relationship.si & sibling\_of \\
ns:film.film.starring/ns:film.performance.actor & starred \\
ns:film.film.written\_by & written\_by \\
ns:film.writer.film & wrote \\
ns:film.actor & actor \\
ns:film.film\_art\_director & art\_director \\
ns:film.cinematographer & cinematographer \\
ns:film.cinematographer.film & cinematographer\_of \\
ns:film.film\_costumer\_designer & costume\_designer \\
ns:film.director & film\_director \\
ns:film.editor & film\_editor \\
ns:business.employer & employer \\
ns:fictional\_universe.fictional\_character & fictional\_character \\
ns:film.film & film \\
ns:film.film\_distributor & film\_distributor \\
ns:people.person & person \\
ns:film.producer & film\_producer \\
ns:film.production\_company & production\_company \\
ns:film.writer & writer \\
ns:m.05zppz & male \\
ns:m.02zsn & female \\
ns:m.0f8l9c & French \\
ns:m.06mkj & Spanish \\
ns:m.0b90\_r & Mexican \\
ns:m.03rjj & Italian \\
ns:m.0d0vqn & Swedish \\
ns:m.09c7w0 & American \\
ns:m.0d060g & Canadian \\
ns:m.0345h & German \\
ns:m.03\_3d & Japanese \\
ns:m.07ssc & British \\
ns:m.059j2 & Dutch \\
ns:m.0d05w3 & Chinese \\

    \end{tabular}
    \caption{Mapping from Freebase identifiers (truncated to first 60 characters) to shorter, more readable strings.}
    \label{tab:mapping}
\end{table*}

\end{document}